\documentclass[11pt,a4paper]{article}
\usepackage[hyperref]{acl2021}
\usepackage{times}
\usepackage{latexsym}

\usepackage{microtype}

\aclfinalcopy 
\def\aclpaperid{2468} 

\usepackage{graphicx} 
\usepackage{amsmath} 
\usepackage{amsfonts}
\usepackage{multirow}
\usepackage{booktabs}
\usepackage[ruled]{algorithm2e}

\title{Adjacency List Oriented Relational Fact Extraction \\via Adaptive Multi-task Learning}

\author{Fubang Zhao$^1$\thanks{~~These two authors contributed equally to this research.}~, Zhuoren Jiang$^2$\footnotemark[1]~\thanks{~~Zhuoren Jiang is the corresponding author}~, Yangyang Kang$^1$, Changlong Sun$^1$, ~Xiaozhong Liu$^3$  \\
$^1$Alibaba Group, Hangzhou, China\\
$^2$School of Public Affairs, Zhejiang University, Hangzhou, China\\
$^3$School of Informatics, Computing and Engineering, IUB, Bloomington, USA\\
{\tt fubang.zfb@alibaba-inc.com, jiangzhuoren@zju.edu.cn}\\
{\tt yangyang.kangyy@alibaba-inc.com, changlong.scl@taobao.com}\\
{\tt liu237@indiana.edu}}

\date{}

\begin{document}
\maketitle
\begin{abstract}
Relational fact extraction aims to extract semantic triplets from unstructured text. In this work, we show that all of the relational fact extraction models can be organized according to a graph-oriented analytical perspective. An efficient model, a\textbf{D}jacency l\textbf{I}st o\textbf{R}iented r\textbf{E}lational fa\textbf{C}T (\textbf{DIRECT}), is proposed based on this analytical framework. To alleviate challenges of error propagation and sub-task loss equilibrium, DIRECT employs a novel adaptive multi-task learning strategy with dynamic sub-task loss balancing. Extensive experiments are conducted on two benchmark datasets, and results prove that the proposed model outperforms a series of state-of-the-art (SoTA) models for relational triplet extraction.
\end{abstract}

\section{Introduction}\label{sec:intro}
Relational fact extraction, as an essential NLP task, is playing an increasingly important role in knowledge graph construction~\cite{han2019opennre,distiawan2019neural}. It aims to extract relational triplet from the text. A relational triplet is in the form of $(subject, relation, object)$ or $(s, r, o)$~\cite{zeng2019learning}. While various prior models proposed for relational fact extraction, few of them analyze this task from the perspective of output data structure.

\begin{figure}[ht]
\centering
\includegraphics[width=1.0\columnwidth]{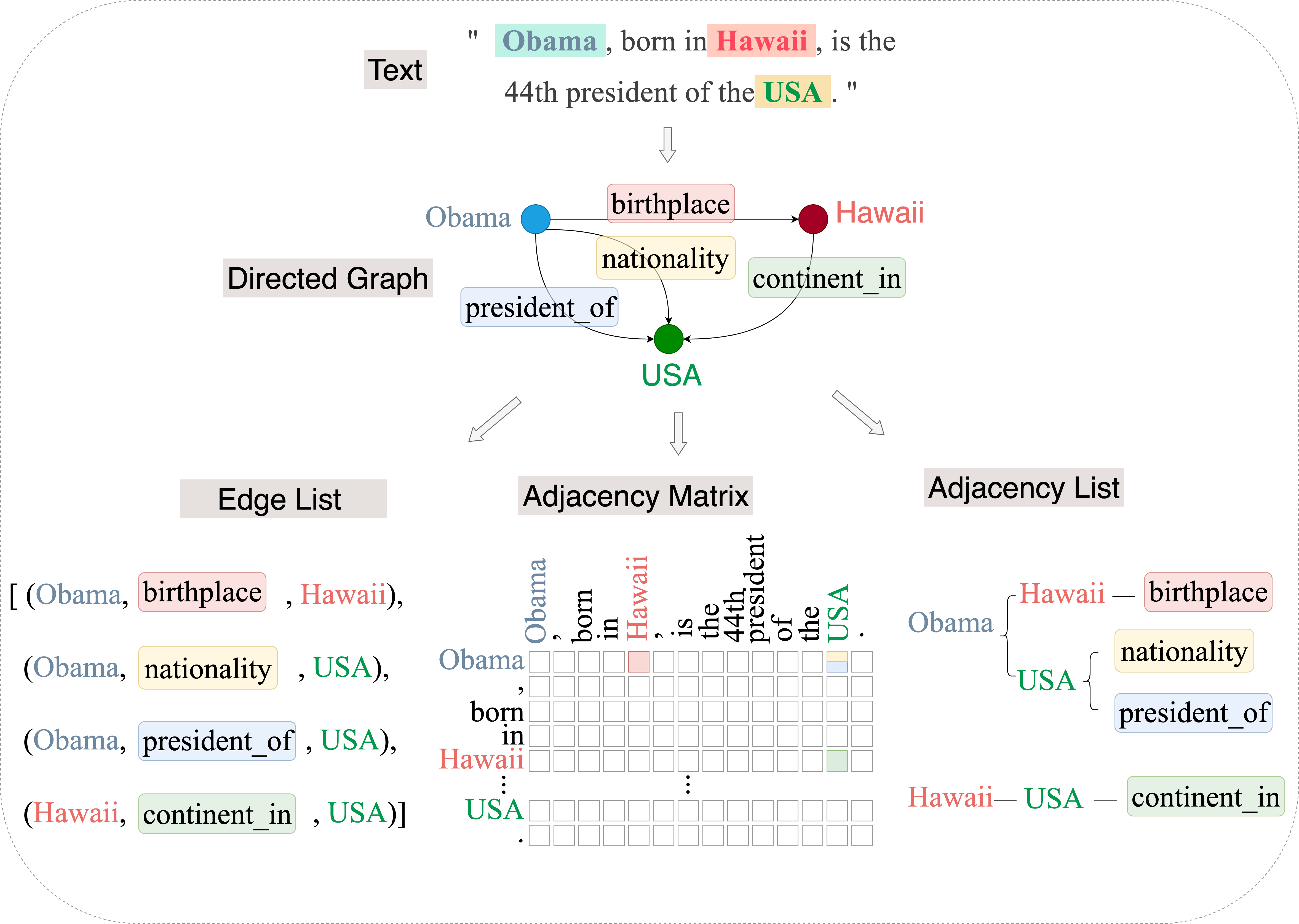}
\caption{Example of exploring the relational fact extraction task from the perspective of directed graph representation method as output data structure.}
\label{fig:example}
\end{figure}

As shown in Figure~\ref{fig:example}, the relational fact extraction can be characterized as a directed graph construction task, where graph representation flexibility and heterogeneity accompany additional benefaction. In practice, there are three common ways to represent graphs~\cite{gross2005graph}: 

\textbf{Edge List} is utilized to predict a sequence of triplets (edges). The recent sequence-to-sequence based models, such as NovelTagging~\cite{zheng2017joint}, CopyRE~\cite{zeng2018extracting}, CopyRL~\cite{zeng2019learning}, and PNDec~\cite{nayak2020effective}, fall into this category. 

Edge list is a simple and space-efficient way to represent a graph~\cite{arifuzzaman2015fast}. However, there are three problems. First, the triplet overlapping problem~\cite{zeng2018extracting}. For instance, as shown in Figure~\ref{fig:example}, for triplets (Obama, nationality, USA) and (Obama, president\_of, USA), there are two types of relations between the ``Obama'' and ``USA''. If the model only generates one sequence from the text~\cite{zheng2017joint}, it may fail to identify the multi-relation between entities. Second, to overcome the triplet overlapping problem, the model may have to extract the triplet element repeatedly~\cite{zeng2018extracting}, which will increase the extraction cost. Third, there could be an ordering problem~\cite{zeng2019learning}: for multiple triplets, the extraction order could influence the model performance.

\textbf{Adjacency Matrices} are used to predict matrices that represent exactly which entities (vertices) have semantic relations (edges) between them. Most early works, which take a pipeline approach~\cite{zelenko2003kernel,zhou2005exploring}, belong to this category. These models first recognize all entities in text and then perform relation classification for each entity pair. The subsequent neural network-based models~\cite{bekoulis2018joint,dai2019joint}, that attempt to extract entities and relations jointly, can also be classified into this category. 

Compared to edge list, adjacency matrices have better relation (edge) searching efficiency~\cite{arifuzzaman2015fast}. Furthermore, adjacency matrices oriented models is able to cover different overlapping cases~\cite{zeng2018extracting} for relational fact extraction task. But the space cost of this approach can be expensive. For most cases, the output matrices are very sparse. For instance, for a sentence with $n$ tokens, if there are $m$ kinds of relations, the output space is $n \cdot n \cdot m$, which can be costly for graph representation efficiency. This phenomenon is also illustrated in Figure~\ref{fig:example}.

\textbf{Adjacency List} is designed to predict  
an array of linked lists that serves as a representation of a graph. As depicted in Figure~\ref{fig:example}, in the adjacency list, each vertex $v$ (key) points to a list (value) containing all other vertices connected to $v$ by several edges. Adjacency list is a hybrid graph representation between edge list and adjacency matrices~\cite{gross2005graph}, which can balance space and searching efficiency\footnote{More detailed complexity analyses of different graph representations are provided in Appendix section~\ref{ass:cagr}.}. Due to the structural characteristic of the adjacency list, this type of model usually adopts a cascade fashion to identify subject, object, and relation sequentially. For instance, the recent state-of-the-art model CasRel~\cite{wei2020novel} can be considered as an exemplar. It utilizes a two-step framework to recognize the possible object(s) of a given subject under a specific relation. However, CasRel is not fully adjacency list oriented: in the first step, it use $subject$ as the key; while in the second step, it predicts $(relation, object)$ pairs using adjacency matrix representation.

Despite its considerable potential, the cascade fashion of adjacency list oriented model may cause problems of sub-task error propagation~\cite{shen2019multi}, i.e., errors from ancestor sub-tasks may accumulate to threaten downstream ones, and sub-tasks can hardly share supervision signals. Multi-task learning~\cite{caruana1997multitask} can alleviate this problem, however, the sub-task loss balancing problem~\cite{chen2018gradnorm,sener2018multi} could compromise its performance.

Based on the analysis from the perspective of output data structure, we propose a novel solution, a\textbf{D}jacency l\textbf{I}st o\textbf{R}iented r\textbf{E}lational fa\textbf{C}T extraction model (\textbf{DIRECT}), with the following advantages:

$\bullet$ For efficiency, DIRECT is a fully adjacency list oriented model, which consists of a shared BERT encoder, the Pointer-Network based subject and object extractors, and a relation classification module. In Section~\ref{sec:computation}, we provide a detailed comparative analysis\footnote{Theoretical representation efficiency analysis of graph representative models are described in Appendix section~\ref{ass:gre}.} to demonstrate the efficiency of the proposed method.

$\bullet$ From the performance viewpoint, to address sub-task error propagation and sub-task loss balancing problems, DIRECT employs a novel adaptive multi-task learning strategy with the dynamic sub-task loss balancing approach. In Section \ref{sec:result} and \ref{sec:aba}, the empirical experimental results demonstrate DIRECT can achieve the state-of-the-art performance of relational fact extraction task, and the adaptive multi-task learning strategy did play a positive role in improving the task performance.

The major contributions of this paper can be summarized as follows: 

1. We refurbish the relational fact extraction problem by leveraging an 
analytical framework of graph-oriented output structure. To the best of our knowledge, this is a pioneer investigation to explore the output data structure of relational fact extractions.

2. We propose a novel solution, DIRECT\footnote{To help other scholars reproduce the experiment outcome, we will release the code and datasets via GitHub: https://github.com/fyubang/direct-ie.\label{ft:code}}, which is a fully adjacency list oriented model with a novel adaptive multi-task learning strategy.

3. Through extensive experiments on two benchmark datasets\textsuperscript{\ref{ft:code}}, we demonstrate the efficiency and efficacy of DIRECT. The proposed DIRECT outperforms the state-of-the-art baseline models.
\section{The DIRECT Framework}\label{sec:method}

\begin{figure*}[ht]
\centering
\includegraphics[width=16cm]{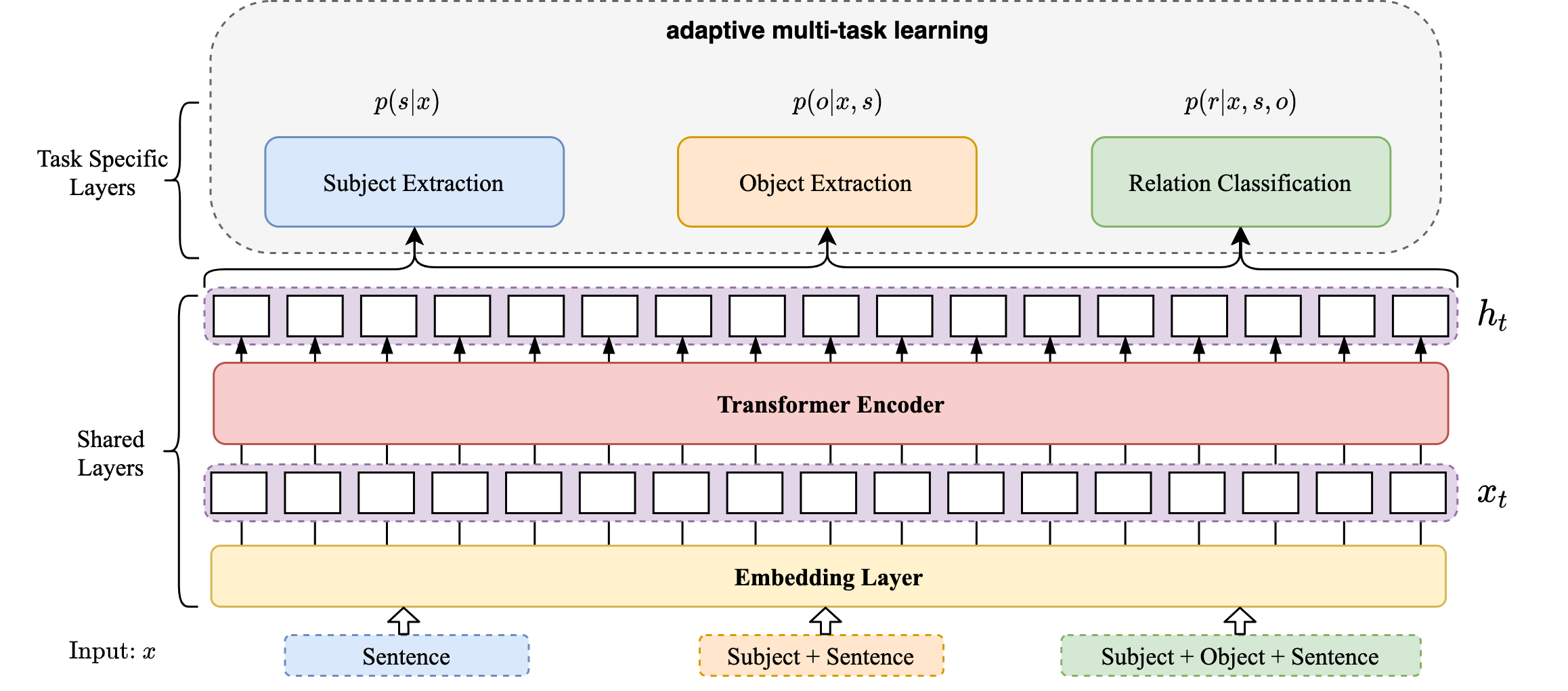}
\caption{An overview of the proposed DIRECT framework}
\label{fig:model}
\end{figure*}
In this section, we will introduce the framework of the proposed DIRECT model, which includes a shared BERT encoder and three output layers: subject extraction, object extraction, and relation classification. As shown in Figure \ref{fig:model}, DIRECT is fully adjacency list oriented. The input sentence is firstly fed into the subject extraction module to extract all subjects. Then each extracted subject is concatenated with the sentence, and fed into the object extraction module to extract all objects, which can form a set of subject-object pairs. Finally, the subject-object pair is concatenated with sentence, and fed into the relation classification module to get the relations between them. For balancing the weights of sub-task losses and to improve the global task performance, three modules share the BERT encoder layer and are trained with an adaptive multi-task learning strategy. 

\subsection{Shared BERT Encoder}
In the DIRECT framework, the encoder is used to extract the semantic features from the inputs for three modules. As aforementioned, we employ the BERT~\cite{devlin2019bert} as the shared encoder to make use of its pre-trained knowledge and attention mechanism. 

The architecture of the shared method is shown in Figure \ref{fig:model}. The lower embedding layer and transformers~\cite{vaswani2017attention} are shared across all the three modules, while the top layers represent the task-speciﬁc outputs.

The encoding process is as follows:
\begin{equation}
    \mathbf{h}^t = \text{BERT}(\mathbf{x}^t)
\end{equation}
where $\mathbf{x}^t = [w_1, ..., w_n]$ is the input text of task $t$ and $\mathbf{h}^t$ is the hidden vector sequence of the input. Due to the limited space, the detailed architecture of BERT please refer to the original paper~\cite{devlin2019bert}.

\subsection{Subject and Object Extraction}
The subject and object extraction modules are motivated by the Pointer-Network~\cite{vinyals2015pointer} architecture, which are 
widely used in Machine Reading Comprehension (MRC) \cite{rajpurkar2016squad} task. Different from MRC task that only needs to extract a single span, the subject and object extractions need to extract multiple spans. Therefore, in the training phase, we replace $softmax$ function with $sigmoid$ function for the activation function of the output layer, and replace cross entropy (CE)~\cite{goodfellow2016deep} with binary cross entropy (BCE)~\cite{luc2016semantic} for the loss function. Specifically, we will perform independent binary classifications for each token twice to indicate whether the current token is the start or the end of a span. The probability of a token to be start or end is as follows:
\begin{align}
    p_{i,{\text{start}}}^t &= \sigma(\mathbf{W}_{\text{start}}^t \cdot h_i + \mathbf{b}_{\text{start}}^t) \\
    p_{i,{\text{end}}}^t &= \sigma(\mathbf{W}_{\text{end}}^t \cdot h_i + \mathbf{b}_\text{end}^t)
\end{align}
where $h_i$ represents the hidden vector of the $i_{th}$ token, $t \in [s, o]$ represents subject and object extraction respectively, $\mathbf{W}^{t} \in \mathbb{R}^{h \times 1}$ represents the trainable weight, $\mathbf{b}^{t} \in \mathbb{R}^1$ is the bias and $\sigma$ is $sigmoid$ function.

During inference, we first recognize all the start positions by checking if the probability $p_{i,{\text{start}}}^t > \alpha$, where $\alpha$ is the threshold of extraction. Then, we identify the corresponding end position with the largest probability $p_{i,{\text{end}}}^t$ between two neighboring start positions. Concretely, assuming $\text{pos}_{j, \text{start}}$ is the start position of the $j_{th}$ span, the corresponding end position is:

\begin{equation}
    \text{pos}_{j, \text{end}} = \underset{\text{pos}_{j, \text{start}} <= i < \text{pos}_{j+1, \text{start}}}{\mathrm{argmax}} p_{i,{\text{end}}}^t
\end{equation}

Though the overall structure is similar, the inputs for subject and object extraction are different. When extracting the subject, only the original sentence needs to be input:
\begin{align}
    \mathbf{x} &= [w_1, ..., w_n] \\
    \text{input}^s &= [\text{[cls]}, \mathbf{x}, \text{[sep]}]
\end{align}
where $w_i$ represents the $i_{th}$ token of the original sentence.

Meanwhile, the object extraction is based on the corresponding subject. To form the input, the subject $s$ and the original sentence $x$ are concatenated with $[sep]$ as follows:
\begin{equation}
    \text{input}^o = [\text{[cls]}, s, \text{[sep]}, \mathbf{x}, \text{[sep]}]
\end{equation}

\subsection{Relation classification}
The output layer of relation classification is relatively simple, which is a normal multi-label classification model. The $[cls]$ vector obtained by BERT encoder is used as the sentence embedding. A fully connected layer is used for the nonlinear transformation, and perform multi-label classification to predict relations of the input subject-object pair. The detailed operations of relation classification are as follows:

\begin{equation}
    \mathbf{P}^{r} = \sigma(\mathbf{W}^{r} \cdot h_{\text{[cls]}} + \mathbf{b}^{r}) 
\end{equation}

where $\mathbf{P}^{r} \in \mathbb{R}^c$ is the predicted probability vector of relations, $\sigma$ is $sigmoid$ function, $\mathbf{W}^{r} \in \mathbb{R}^{h \times c}$ and $\mathbf{b}^{r} \in \mathbb{R}^c$ are the trainable weights and bias, $h$ is the hidden size of encoder, $c$ is the number of relations, and $h_{\text{[cls]}}$ denotes the hidden vector of the first token $\text{[cls]}$. The input for relation classification task is as follows:
\begin{equation}
    \text{input}^r = [\text{[cls]}, s, \text{[sep]}, o, \text{[sep]}, \mathbf{x}, \text{[sep]}]
\end{equation}

\begin{algorithm}[ht]
\caption{Adaptive Multi-task Learning with Dynamic Loss Balancing}
\label{algo:mtl}
\LinesNumbered 
Initialize model parameters $\Theta$ randomly\;
Load pre-trained BERT parameters for shared encoder\;
Prepare the data for each task $t$ and pack them into mini-batch: $D^t$, $t\in [s, o, r]$ \;
Get the number of batch for each task: $n^t$\; 
Set the number of epoch for training: $epoch_{max}$\;
\For{ $epoch$ in $1, 2, ..., epoch_{max}$}{
    1. Merge all the datasets: $D = D^{s} \cup D^{o} \cup D^{r}$\;
    2. Shuffle D\;
    3. Initialize EMA for each task $v^t=1$ and its decay $\epsilon=0.99$ \;
    \For{$b^t$ in  $D$}{
        // $b^t$ is a mini-batch of $D^t$ \;
        4. Compute loss: $l^t(\Theta)$ \;
        5. Update EMA: $v^t = (1-\epsilon)\cdot\sum(l^t)+\epsilon \cdot v^t$ \;
        6. Calculate and normalize the weights: $w^t = (v^t/n^t) / (v^{r}/n^{r})$ \;
        7. Update model $\Theta$  with gradient: $\nabla(w^t\cdot \bar{l^t})$ \;
    }
}
\end{algorithm}

\subsection{Adaptive Multi-task Learning}
In DIRECT, subject extraction module, object extraction module, and relation classification module can be considered as three sub-tasks. As aforementioned, if we train each module directly and separately, the error propagation problem would reduce the task performance. Meanwhile, three independent encoders would consume more memory. Therefore, we use multi-task learning to alleviate this problem, and the encoder layer is shared across three modules.

However, applying multi-task learning could be challenging in DIRECT, due to the following problems:

$\bullet$ The input and output of the three modules are different, which means we cannot simply sum up the loss of each task.

$\bullet$ How should we balance the weights of losses for three sub-task modules?

These issues can affect the final results of multi-task training~\cite{shen2019multi,sener2018multi}.

In this work, based on the architecture of MT-DNN~\cite{liu2019multi}, we propose a novel adaptive multi-task learning strategy to address the above problems. The algorithm is shown as Algorithm \ref{algo:mtl}. Basically, the datasets are firstly split into mini-batches. A batch is then randomly sampled to calculate the loss. The parameters of the shared encoder and its task-specific layer are updated accordingly. Especially, the learning effect of each task $t$ is different and dynamically changing during training. Therefore, an approach of adaptively adjusting the weights of task losses is applied. The sum of sub-task's loss $\sum l^t$ is utilized to approximate its optimization effect. The adaptive weight adjusting strategy ensures that the more room a sub-task has to be optimized, the more weight its loss will receive. Furthermore, an exponential moving average (EMA) \cite{lawrance1977exponential} is maintained to avoid the drastic fluctuations of loss weights. Last but not least, to make sure that each task has enough influence on the shared encoder, the weight of the sub-task will be penalized according to the training data amount of each sub-task.


\section{Experiments}
\subsection{Dataset and Experiment Setting}
\textbf{Datasets}. Two public datasets are used for evaluation: \textbf{NYT}~\cite{riedel2010modeling} is originally produced by the distant supervision approach. There are 1.18M sentences with 24 predeﬁned relation types in NYT. \textbf{WebNLG}~\cite{gardent2017creating} is originally created for Natural Language Generation (NLG) tasks. \cite{zeng2018extracting} adopts this dataset for relational triplet extraction task. It contains 246 predeﬁned relation types. There are different versions of these two datasets. To facilitate comparison evaluation, we use the datasets released by~\cite{zeng2018extracting} and follow their data split rules. 

Besides the basic relational triplet extraction, recent studies are focusing on the relational triplet overlapping problem~\cite{zeng2018extracting,wei2020novel}. Follow the overlapping pattern definition of relational triplets~\cite{zeng2018extracting}, the sentences in both datasets are divided into three categories, namely, Normal, EntityPairOverlap (EPO), and SingleEntityOverlap (SEO). The statistics of the two datasets are described in Table~\ref{table:dataset}. 
   
\begin{table}[htbp]
\centering
\begin{tabular}{ccccc}
\toprule[1pt]
\multirow{2}*{Category} & \multicolumn{2}{c}{NYT} &\multicolumn{2}{c}{WebNLG}\\
\cline{2-3} \cline{4-5}
&Train&Test&Train&Test\\
\hline
Normal&37013&3266&1596&246 \\
EPO&9782&978&227&26 \\
SEO&14735&1297&3406&457 \\
\hline
ALL&56195&5000&5019&703 \\
\bottomrule[1pt]
\end{tabular}
\caption{Statistics of Dataset NYT and WebNLG. Note that a sentence can belong to both EPO class and SEO class.}
\label{table:dataset}
\end{table}

\textbf{Baselines}: the following strong state-of-the-art (SoTA) models have been compared in the experiments. 

$\bullet$ NovelTagging~\cite{zheng2017joint} introduces a tagging scheme that transforms the joint entity and relation extraction task into a sequence labeling problem. It can be considered as edge list oriented. 

$\bullet$ CopyRE~\cite{zeng2018extracting} is a seq2seq based model with the copy mechanism, which can effectively extract overlapping triplets. It has two variants: CopyRE$_\text{one}$ employs one decoder; CopyRE$_\text{mul}$ employs multiple decoders. CopyRE is also edge list oriented.

$\bullet$ GraphRel~\cite{fu2019graphrel} is a GCN (graph convolutional networks)~\cite{kipf2017semi} based model, where a relation-weighted GCN is utilized to learn the interaction between entities and relations. It is a two phases model: GraphRel$_\text{1p}$ denotes 1st-phase extraction model; GraphRel$_\text{2p}$ denotes full extraction model. GraphRel is adjacency matrices oriented. 

$\bullet$ CopyRL~\cite{zeng2019learning} combines the reinforcement learning with a seq2seq model to automatically learn the extraction order of triplets. CopyRL is edge list oriented. 

$\bullet$ CasRel~\cite{wei2020novel} is a cascade binary tagging framework, where all possible subjects are identiﬁed in the ﬁrst stage, and then for each identiﬁed subject, all possible relations and the corresponding objects are simultaneously identiﬁed by a relation speciﬁc tagger. This work recently achieves the SoTA results. As aforementioned, CasRel is partially adjacency list oriented.

\textbf{Evaluation Metrics}: following the previous work~\cite{zeng2018extracting,wei2020novel}, different models are compared by using standard micro Precision (Prec.), Recall (Rec.), and F1-score\footnote{In this study, the results of baseline models are all self-reported results from their original papers. Meanwhile, the experimental results of our proposed model are the average of five runs.}. An extracted relational triplet (subject, relation, object) is regarded as correct only if the relation and the heads of both subject and object are all correct.

\textbf{Implementation Details}. The hyper-parameters are determined on the validation set. To avoid the evaluation bias, all reported results from our method are averaged results for 5 runs. More implementation details are described in Appendix section~\ref{ass:id}.


\begin{table*}[htbp]
\centering
\small
\begin{tabular}{c|c|ccc|ccc}
\toprule[1pt]
\multirow{2}*{Method} & \multirow{2}*{Category} & \multicolumn{3}{c|}{NYT} &\multicolumn{3}{c}{WebNLG}\\
\cline{3-5} \cline{6-8}
&&Prec.&Rec.&F1&Prec.&Rec.&F1\\
\hline
$\text{NovelTagging}$\cite{zheng2017joint}& EL & 62.4&31.7&42.0&52.5&193.&28.3\\
$\text{CopyRE}_\text{One}$\cite{zeng2018extracting}& EL&59.4&53.1&56.0&32.2&28.9&30.5\\
$\text{CopyRE}_\text{Mul}$\cite{zeng2018extracting}&EL&61.0&56.6&58.7&37.7&36.4&37.1 \\
$\text{GraphRel}_\text{1p}$\cite{fu2019graphrel}&AM&62.9&57.3&60.0&42.3&39.2&40.7 \\
$\text{GraphRel}_\text{2p}$\cite{fu2019graphrel}&AM&63.9&60.0&61.9&44.7&41.1&42.9 \\
$\text{CopyRL}$\cite{zeng2019learning}&EL&77.9&67.2&72.1&63.3&59.9&61.6 \\
$\text{CasRel}$\cite{wei2020novel}&AL$_\text{P}$&89.7&89.5&89.6&93.4&90.1&91.8 \\
\hline
$\text{DIRECT(Ours)}$ & AL$_\text{F}$& \textbf{92.3}$\pm$\textbf{0.32}& \textbf{92.8}$\pm$\textbf{0.26}&\textbf{92.5}$\pm$\textbf{0.09}& \textbf{93.6}$\pm$\textbf{0.1}&\textbf{92.7}$\pm$\textbf{0.24}& \textbf{93.2}$\pm$\textbf{0.07}\\
\bottomrule[1pt]
\end{tabular}
\caption{Results of different methods on NYT and WebNLG datasets. EL: Edge List; AM: Adjacency Matrices; AL$_\text{P}$: Adjacency List (Partially); AL$_\text{F}$: Adjacency List (Fully).}
\label{table:res_main}
\end{table*}

\subsection{Results and Analysis}\label{sec:result}
\textbf{Relational Triplet Extraction Performance}. The task performances on two datasets are summarized in Table~\ref{table:res_main}. Based on the experiment results, we have the following observations and discussions:

$\bullet$ The proposed DIRECT model outperformed all baseline models in terms of all evaluation metrics on both datasets, which proved DIRECT model can effectively address the relational triplet extraction task.

$\bullet$ The best-performed model (DIRECT) and runner-up model (CasRel) were both adjacency list oriented model. These two models overwhelmingly outperformed other models, which indicated the considerable potential of adjacency list (as the output data structure) for improving the task performance.

\begin{table}[htbp]
\centering
\begin{tabular}{c|c|c|c}
\toprule[1pt]
{Method} &{Element} & NYT & WebNLG\\
\hline
\multirow{3}*{CasRel}&s&93.5 &95.7 \\
&o&93.5 &95.3 \\
&r&94.9&94.0 \\
\hline
\multirow{3}*{DIRECT(Ours)}&s&\textbf{95.4}&\textbf{97.3} \\
&o&\textbf{96.4}&\textbf{96.4} \\
&r&\textbf{97.8}&\textbf{97.4} \\
\bottomrule[1pt]
\end{tabular}
\caption{F1-score for extracting elements of relational triplets on NYT and WebNLG datasets.}
\label{table:res_rte}
\end{table}

$\bullet$ To further compare the relation extraction ability of DIRECT and CasRel, we took a closer look at the extraction performance of relational triplet elements from these two models. As shown in Table~\ref{table:res_rte}\footnote{More detailed results with Precision and Recall are provided in Appendix section~\ref{ass:ser}.}, DIRECT outperformed CasRel in terms of all relational triplet elements on both datasets. These empirical results suggested that, for relational triplet extraction, a fully adjacency list oriented model (DIRECT) may have advantages over a partially oriented one (CasRel).

\begin{figure}[h]
\centering
\includegraphics[width=1.0\columnwidth]{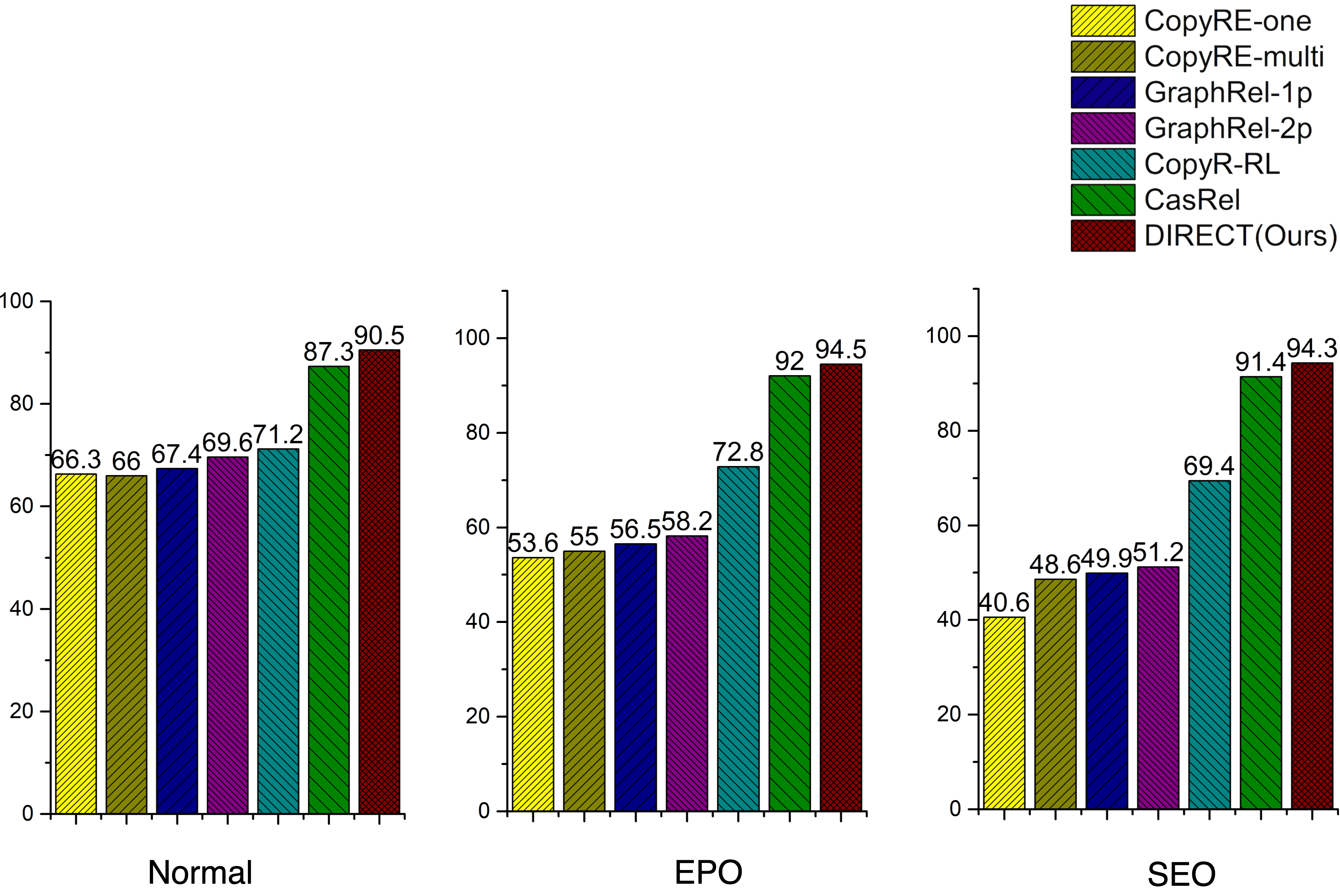}
\caption{F1 score of extracting relational triples from sentences with different overlapping patterns on NYT dataset.}
\label{fig:overlap}
\end{figure}

\textbf{Ability in Handling The Overlapping Problem}. The relational facts in sentences are often complicated. Different relational triplets may have overlaps in a sentence. To verify the ability of our models in handling the overlapping problem, we conducted further experiments on NYT dataset. Figure~\ref{fig:overlap} illustrated of F1 scores of extracting relational triplets from sentences with different overlapping patterns. DIRECT outperformed all baseline models in terms of all overlapping patterns. These results demonstrated the effectiveness of the proposed model in solving the overlapping problem.

\textbf{Ability in Handling Multiple Relation Extraction}. We further compared the model’s ability of extracting relations from sentences that contain multiple triplets. The sentences in NYT and WebNLG were divided into 5 categories. Each category contained sentences that had 1,2,3,4 or $\geq$ 5 triplets. The triplet number was denoted as $N$. As shown in Table~\ref{table:res_dnt}:

$\bullet$ DIRECT achieved the best performance for all triplet categories on both datasets.  These experimental results demonstrated our model had an excellent ability in handling multiple relation extraction.

$\bullet$ In both NYT and WebNLG datasets, when the sentences contained more triplets, the leading advantage of DIRECT became greater. This observation indicated that DIRECT was good at solving complex relational fact extraction.

\begin{table*}[htbp]
\centering
\small
\begin{tabular}{c|ccccc|ccccc}
\toprule[1pt]
\multirow{2}*{Method} & \multicolumn{5}{c|}{NYT} &\multicolumn{5}{c}{WebNLG}\\
\cline{2-6} \cline{7-11}
&$N=1$&$N=2$&$N=3$&$N=4$&$N\ge5$&$N=1$&$N=2$&$N=3$&$N=4$&$N\ge5$\\
\hline
Count&3244&1045&312&291&108&268&174&128&89&44 \\
\hline
$\text{CopyRE}_\text{One}$&66.6&52.6&49.7&48.7&20.3&65.2 &33.0&22.2&14.2&13.2 \\
$\text{CopyRE}_\text{Mul}$&67.1&58.6&52.0&53.6&30.0&59.2&42.5&31.7&24.2&30.0 \\
$\text{GraphRel}_\text{1p}$&69.1&59.5&54.4&53.9&37.5&63.8&46.3&34.7&30.8&29.4 \\
$\text{GraphRel}_\text{2p}$&71.0&61.5&57.4&55.1&41.1&66.0&48.3&37.0&32.1&32.1 \\
$\text{CopyRL}$&71.7&72.6&72.5&77.9&45.9&63.4&62.2&64.4&57.2&55.7 \\
$\text{CasRel}$&88.2&90.3&91.9&94.2&83.7&89.3&90.8&94.2&92.4&90.9 \\
\hline
$\text{DIRECT(Ours)}$&\textbf{90.4}&\textbf{93.1}&\textbf{94.3}&\textbf{95.8}&\textbf{93.1}&\textbf{90.3}&\textbf{92.8}&\textbf{94.8}&\textbf{94.0}&\textbf{92.9} \\
\bottomrule[1pt]
\end{tabular}
\caption{F1-score of extracting relational triplets from sentences with different number (denoted as N) of triplets.}
\label{table:res_dnt}
\end{table*}

\subsection{Ablation Study}\label{sec:aba}
To validate the effectiveness of components in DIRECT, We implemented several model variants for ablation tests\footnote{Due to the length limitation, we list two main ablation experiments, the rest will be provided in the Appendix section \ref{ass:ser}.}. The results of the comparison on NYT dataset are shown in Table~\ref{table:aba}. In particular, we aim to address the following two research questions: 

\textbf{RQ1}: Is it possible to improve the model performance by sharing the parameters of extraction layers?

\textbf{RQ2}: Did the proposed adaptive multi-task learning strategy improve the task performance?

\begin{table}[htbp]
\centering
\begin{tabular}{c|ccc}
\toprule[1pt]
\multirow{2}*{Method} & \multicolumn{3}{c}{NYT} \\
\cline{2-4}
&Prec.&Rec.&F1\\
\hline
DIRECT$_\text{shared}$ & 92.1 & 91.6 & 91.9\\
DIRECT$_\text{equal}$ & 90.6 & 91.3 & 91.0 \\
DIRECT & \textbf{92.3} & \textbf{92.8} & \textbf{92.5}\\
\bottomrule[1pt]
\end{tabular}
\caption{Results of model variants for ablation tests.}
\label{table:aba}
\end{table}
\textbf{Effects of Sharing Extraction Layer Parameters (RQ1)}. As described in Section~\ref{sec:method}, the structures of subject extraction and object extraction output layers are exactly the same. To answer RQ1, we merged the subject extraction and object extraction layers into one entity extraction layer by sharing the parameters of output layers of these two modules, denoted as \textbf{DIRECT$_\mathbf{shared}$}. From the results of Table~\ref{table:aba}, we can observe that, sharing the parameters of output layers of two extraction modules would reduce the performance of the model. A possible explanation is that, although the output of these two modules is similar, the semantics of subject and object are different. Hence, directly sharing the output parameters of two modules could lead to an unsatisfactory performance.

\textbf{Effects of Adaptive Multi-task Learning (RQ2)}. As described in Section~\ref{sec:method}, the adaptive multi-task learning strategy with the dynamic sub-task loss balancing approach is proposed for improving the task performance. To answer RQ2, we replaced the adaptive multi-task learning strategy with an ordinary learning strategy. In this strategy, the losses of three sub-tasks were computed with equal weights, denoted as \textbf{DIRECT$_\mathbf{equal}$}. From the results of Table~\ref{table:aba}, we can observe that, by using adaptive multi-task learning, DIRECT was able to get a 1.5 percentage improvement on the F1-score. This significant improvement indicated that adaptive multi-task learning played a positive role in the balance of sub-task learning and can improve the global task performance.



\begin{table}[htbp]
\centering
\begin{tabular}{c|c|cc}
\toprule[1pt]
Method & Category  & NYT & WebNLG\\
\hline
CopyRe & EL &  329 & 712 \\
MHS & AM  & 57369 & 26518 \\
CasRel & AL$_\text{P}$ & 3084 & 15836 \\
\hline
DIRECT& AL$_\text{F}$ &  \textbf{238} & \textbf{542} \\
\bottomrule[1pt]
\end{tabular}
\caption{Graph representation efficiency estimation based on the predicted logits amount. EL: Edge List; AM: Adjacency Matrices; AL$_\text{P}$: Adjacency List (Partially); AL$_\text{F}$: Adjacency List (Fully).}
\label{tab:logit1}
\end{table}

\subsection{Graph Representation Efficiency Analysis}\label{sec:computation}
Based on the amount estimation of predicted logits\footnote{Numeric output $(0/1)$ of the last layer}, we conduct a graph representation efficiency analysis to demonstrate the efficiency of the proposed method\footnote{From the graph representation perspective, when a method requires fewer logits to represent the graph (set of triples), it will reduce the model fitting difficulty.}.

For each graph representation category, we choose one representative  algorithms. \textbf{Edge List}: CopyRE~\cite{zeng2018extracting}; \textbf{Adjacency Matrices}: MHS~\cite{bekoulis2018joint}; \textbf{Adjacency List}: CasRel (partially)~\cite{wei2020novel} and the proposed DIRECT (fully).

The averaged predicted logits estimation for one sample\footnote{The theoretical analysis of predicted logits for different models are described in Appendix section \ref{ass:gre}.} of different models on two datasets are shown in Table~\ref{tab:logit1}. MHS is adjacency matrices oriented, it has the most logits that need to be predicted. Since CasRel is partially adjacency list oriented, it needs to predict more logits than DIRECT. Theoretically, as an edge list oriented, the predicted logits of CopyRE should be the least. But, as described in Section~\ref{sec:intro}, it needs to extract the entities repeatedly to handle the overlapping problem. Hence, its graph representation efficiency could be worse than our model. The structure of our model is simple and fully adjacency list oriented. Therefore, from the viewpoint of predicted logits estimation, DIRECT is the most representative-efficient model.
\section{Related Work}\label{review}
\textbf{Relation Fact Extraction}. In this work, we show that all of the relational fact extraction models can be unified into a graph-oriented output structure analytical framework. From the perspective of graph representation, the prior models can be divided into three categories.
\textbf{Edge List}, this type of model usually employs sequence-to-sequence fashion, such as NovelTagging~\cite{zheng2017joint}, CopyRE~\cite{zeng2018extracting}, CopyRL~\cite{zeng2019learning}, and PNDec~\cite{nayak2020effective}. Some models of this category may suffer from the triplet overlapping problem and expensive extraction cost. \textbf{Adjacency Matrices}, many early pipeline approaches~\cite{zelenko2003kernel,zhou2005exploring,mintz2009distant} and recent neural network-based models~\cite{bekoulis2018joint,dai2019joint,fu2019graphrel}, can be classified into this category. The main problem for this type of model is the graph representation efficiency. \textbf{Adjacency List}, the recent state-of-the-art model CasRel~\cite{wei2020novel} is a partially adjacency list oriented model. In this work, we propose DIRECT that is a fully adjacency list oriented relational fact extraction model. To the best of our knowledge, few previous works analyze this task from the output data structure perspective. GraphRel~\cite{fu2019graphrel} employs a graph-based approach, but it is utilized from an encoding perspective, while we analyze it from the perspective of output structure. Our work is a pioneer investigation to analyze the output data structure of relational fact extraction.

\textbf{Multi-task Learning}. Multi-task Learning (MTL) can improve the model performance. \cite{caruana1997multitask} summarizes the goal succinctly: \textit{``it improves generalization by leveraging the domain-specific information contained in the training signals of related task.''} It has two benefits~\cite{vandenhende2020multi}: (1) multiple tasks share a single model, which can save memory. (2) Associated tasks complement and constrain each other by sharing information, which can reduce overfitting and improve global performance. There are two main types of MTL: hard parameter sharing~\cite{baxter1997bayesian} and soft parameter sharing~\cite{duong2015low}. Most of the multi-task learning is done by summing the loses directly, this approach is not suitable 
for our case. When the input and output are different, it is impossible to get two losses in one forward propagation. MT-DNN~\cite{liu2019multi} is proposed for this problem. Furthermore, MTL is difficult for training, the magnitudes of different task-losses are different, and the direct summation of losses may lead to a bias for a particular task. There are already some studies proposed to address this problem~\cite{chen2018gradnorm,guo2018dynamic,liu2019end}. They all try to dynamically adjust the weight of the loss according to the magnitude of the loss, the difficulty of the problem, the speed of learning, etc. In this study, we adopt MT-DNN's framework, and propose an adaptive multi-task learning strategy that can dynamically adjust the loss weight based on the averaged EMA~\cite{lawrance1977exponential} of the training data amount, task difficulty, etc.

\section{Conclusion}
In this paper,  we introduce a new analytical perspective to organize the relational fact extraction models and propose DIRECT model for this task. Unlike existing methods, DIRECT is fully adjacency list oriented, which employs a novel adaptive multi-task learning strategy with dynamic sub-task loss balancing. Extensive experiments on two public datasets, prove the efficiency and efficacy of the proposed methods.

\section*{Acknowledgments}
We are thankful to the anonymous reviewers for their helpful comments. This work is supported by Alibaba Group through Alibaba Research Fellowship Program, the National Natural Science Foundation of China (61876003), the Key Research and Development Plan of Zhejiang Province (Grant No.2021C03140), the Fundamental Research Funds for the Central Universities, and Guangdong Basic and Applied Basic Research Foundation (2019A1515010837).

\bibliographystyle{acl_natbib}
\bibliography{acl2021}

\clearpage
\section{Appendix}
\subsection{Implementation Details}\label{ass:id}
We adopted the pre-trained BERT model [BERT-Base-Cased]\footnote{Available at: https://storage.googleapis.com/bert\_models/\\2018\_10\_18/cased\_L-12\_H-768\_A-12.zip} as our encoder, where the number of Transformer layers was 12 and the hidden size was 768. The token types of input were always set to 0. 

We used Adam as our optimizer and applied a triangular learning rate schedule as suggested by original BERT paper. In addition, we adopted a lazy mechanism for optimization. Different from the momentum mechanism of ordinary Adam optimizer~\cite{kingma2015adam} that updated the output layer parameters for all tasks, this lazy-Adam mechanism wouldn't update the parameters of non-current tasks. 

The dacay rate $\epsilon$ of EMA was set to 0.99 as default. The max sequence length was 128.

The other hyper-parameters were determined on the validation set. Notably, considering our special decoding strategy, we raised the threshold of extraction to 0.9 to balance the precision and the recall. The threshold of relation classification was set to 0.5 as default. The hyper-parameter setting was listed in Table \ref{table:res_hyperpara_appendix}. 

Our mthod were implemented by Pytorch\footnote{https://pytorch.org/} and run on a server configured with a Tesla V100 GPU, 16 CPU, and 64G memory.

\begin{table}[htbp]
\centering
\begin{tabular}{c|ccc}
\toprule[1pt]
Hyper-parameter & NYT & WebNLG \\
\hline
Learning Rate & 8e-5 & 1e-4 \\
Epoch Num. & 15 & 60 \\
Batch Size & 32 & 16 \\
\bottomrule[1pt]
\end{tabular}
\caption{Hyper-parameter setting for NYT and WebNLG datasets. }
\label{table:res_hyperpara_appendix}
\end{table}

\subsection{Supplementary Experimental Results}\label{ass:ser}

\textbf{Ablation Study}. To validate the effectiveness of components in DIRECT, We implemented several model variants for ablation tests respectively. For experimental fairness, we kept the other components in the same settings when modifying one module.
\begin{itemize}
    \item \textbf{DIRECT$_\mathbf{shared}$}, we merged the subject extraction and object extraction layers into one entity extraction layer by sharing the parameters of output layers of these two modules. 
    \item \textbf{DIRECT$_\mathbf{equal}$}, we replaced the adaptive multi-task learning strategy with an ordinary learning strategy. In this strategy, the losses of three sub-tasks were computed with equal weights, denoted as \textbf{DIRECT$_\mathbf{equal}$}.
    \item \textbf{DIRECT$_\mathbf{threshold}$}, we simply recognized all the start and end positions of entities by checking if the probability $p_{i,{\text{start/end}}}^t > \alpha$, where $\alpha$ was the threshold of extraction. 
    \item \textbf{DIRECT$_\mathbf{adam}$}, we used ordinary Adam as optimizer. 
    
\end{itemize}

\begin{table}[h]
\centering
\begin{tabular}{c|ccc}
\toprule[1pt]
\multirow{2}*{Method} & \multicolumn{3}{c}{NYT} \\
\cline{2-4}
&Prec.&Rec.&F1\\
\hline
DIRECT$_\text{shared}$ & 92.1 & 91.6 & 91.9\\
DIRECT$_\text{equal}$ & 90.6 & 91.3 & 91.0 \\
DIRECT$_\text{threshold}$ & 92.8 & 92.0 & 92.4 \\
DIRECT$_\text{adam}$ & 92.1 & \textbf{92.9} & 92.5 \\
DIRECT & \textbf{92.9} & 92.1 & \textbf{92.5}\\
\bottomrule[1pt]
\end{tabular}
\caption{Results of model variants for ablation tests.}
\label{table:aba_all}
\end{table}

\begin{table*}[htbp]
\centering
\begin{tabular}{c|c|ccc|ccc}
\toprule[1pt]
\multirow{2}*{Method} &\multirow{2}*{Element} & \multicolumn{3}{c|}{NYT} &\multicolumn{3}{c}{WebNLG}\\
\cline{3-5} \cline{6-8}
&&Prec.&Rec.&F1&Prec.&Rec.&F1\\
\hline
\multirow{3}*{CasRel}&s&94.6 &92.4&93.5 &98.7&92.8&95.7 \\
&o&94.1&93.0&93.5 &97.7&93.0&95.3 \\
&r&96.0&93.8&94.9&96.6&91.5&94.0 \\
\hline
\multirow{3}*{Ours}&s&95.1&95.1&\textbf{95.1}&97.1&96.8&\textbf{96.9} \\
&o&97.2&96.3&\textbf{96.7}&96.4&96.3&\textbf{96.3} \\
&r&98.6&98.3&\textbf{98.5}&97.6&97.3&\textbf{97.4} \\
\bottomrule[1pt]
\end{tabular}
\caption{Results on extracting elements of relational triplets}
\label{table:res_rte_all}
\end{table*}

\begin{table*}[htbp]
\centering
\begin{tabular}{c|ccc}
\toprule[1pt]
\multirow{2}*{Method} & \multicolumn{3}{c}{NYT} \\
\cline{2-4}
&Prec.&Rec.&F1\\
\hline
$\text{MHS}^*$ \cite{bekoulis2018joint}&60.7 &58.6 &59.6 \\
$\text{CopyMTL}_{one}$\cite{zeng2020copymtl} &72.7 &69.2 &70.9 \\
$\text{CopyMTL}_{mul}$\cite{zeng2020copymtl} &75.7 &68.7 &72.0 \\
WDec \cite{nayak2020effective}&88.1 &76.1 &81.7 \\
PNDec \cite{nayak2020effective}&80.6 &77.3 &78.9 \\
Seq2UMTree \cite{zhang2020minimize}& 79.1 & 75.1 & 77.1 \\
\hline
DIRECT(ours) & \textbf{90.2} & \textbf{90.2} & \textbf{90.2}\\
\bottomrule[1pt]
\end{tabular}
\caption{Results of different methods under Exact-Match Metrics. * marks results reproduced by official implementation.}
\label{table:res_exact}
\end{table*}

From the results of Table~\ref{table:aba_all}, we can observe that:
\begin{enumerate}
\item Sharing the parameters of output layers of subject and object extraction modules would reduce the performance of the model.
\item Compared to ordinary multi-task learning strategy, by using adaptive multi-task learning, DIRECT was able to get a 1.5 percentage point improvement on F1-score.
\item There would be a slight drop in performance, if we just used a simple threshold policy to recognize the start and end positions of an entity.
\item Despite the difference in precision and recall, there was
 no significant difference between these two optimizers (ordinary-Adam \& lazy-Adam ) for the task.
\end{enumerate} 
 
\textbf{Results on Extracting Elements of Relational Triplets}. The complete extraction performance of relational triplet elements from DIRECT and CaslRel are listed in Table~\ref{table:res_rte_all}. DIRECT outperformed CasRel in terms of all relational triplet elements on both datasets. These empirical results suggest that, for relational triplet extraction, a fully adjacency list oriented model (DIRECT) may have advantages over a partially oriented one (CasRel).

\begin{table*}[htbp]
\centering
\begin{tabular}{ccccc}
\toprule[1pt]
Category & Method & Theoretical & NYT & WebNLG\\
\hline
Edge List & CopyRe & $4kl+kr$ & 329 & 712 \\
Adjacency Matrices & MHS &$llr$& 57369 & 26518 \\
Adjacency List (Partially)& CasRel & $2l+2slr$ & 3084 & 15836 \\
\hline
Adjacency List (Fully)& DIRECT & $2l+2sl+or$ & 238 & 542 \\
\bottomrule[1pt]
\end{tabular}
\caption{Graph representation efficiency based on the theoretical logits amount and the estimated logits amount on two benchmark datasets.}
\label{tab:logit1_all}
\end{table*}

\textbf{Results of Different Methods under Exact-Match Metrics}. In experiment section, we followed the match metric from~\cite{zeng2018extracting}, which only required to match the first token of entity span. Many previous works adopted this match metric~\cite{fu2019graphrel,zeng2019learning,wei2020novel}. 

In fact, our model is capable of extracting the complete entities. Therefore, we collected papers that reported the results of exact-match metrics (requiring to match the complete entity span). The following strong state-of-the-art (SoTA) models have been compared: 

$\bullet$ CopyMTL \cite{zeng2020copymtl} is a multi-task learning framework, where conditional random ﬁeld is used to identify entities, and a seq2seq model is adopted to extract relational triplets.

$\bullet$ WDec \cite{nayak2020effective} fuses a seq2seq model with a new representation scheme, which enables the decoder to generate one word at a and can handle full entity names of different length and overlapping entities.

$\bullet$ PNDec \cite{nayak2020effective} is a modiﬁcation of seq2seq model. Pointer networks are used in the decoding framework to identify the entities in the sentence using their start and end locations.

$\bullet$ Seq2UMTree \cite{zhang2020minimize} is a modiﬁcation of seq2seq model, which employs an unordered-multi-tree decoder to to minimize exposure bias.

The task performances on NYT dataset are summarized in Table~\ref{table:res_exact}. The proposed DIRECT model outperformed all baseline models in terms of all evaluation metrics. This experimental results further confirmed the efficacy of DIRECT for relational fact extraction task. 

\subsection{Complexity Analysis of Graph Representations}\label{ass:cagr}
For a graph $G=(V,E)$, $|V|$ denotes the number of nodes/entities and $|E|$ denotes the number of edges/relations. Suppose there are $m$ kinds of relations, $d(v)$ denotes the number of edges from node $v$.

$\bullet$  Edge List Complexity
    \begin{enumerate}
    	\item[$-$] Space: $O(|E|)$
        \item[$-$] Find all edges/relations from a node: $O(|E|)$
    \end{enumerate}
    
$\bullet$  Adjacency Matrices Complexity
    \begin{enumerate}
    	\item[$-$] Space: $O(|V| \cdot |V| \cdot m)$
        \item[$-$] Find all edges/relations from a node: $O(|V| \cdot m)$
    \end{enumerate}

$\bullet$  Adjacency List Complexity
    \begin{enumerate}
    	\item[$-$] Space: $O(|V|+|E|)$
        \item[$-$] Find all edges/relations from a node: $O(d(v))$
    \end{enumerate}

\subsection{Graph Representation Efficiency Analysis}\label{ass:gre}
Based on the amount estimation of predicted logits\footnote{Numeric output of the last layer} $(0/1)$, we conduct a graph representation efficiency analysis to demonstrate the efficiency of proposed method\footnote{As aforementioned, from the graph representation perspective, when a method requires fewer logits to represent the graph (set of triples), it will reduce the model fitting difficulty.}.

For each graph representation category, we choose one representative model algorithms. \textbf{Edge List}: CopyRE~\cite{zeng2018extracting}; \textbf{Adjacency Matrices}: MHS~\cite{bekoulis2018joint}; \textbf{Adjacency List}: CasRel (partially)~\cite{wei2020novel} and DIRECT (fully).

Formally, for a sentence whose length is $l$ ($l$ tokens), there are  $r$ types of relations, $k$ denotes the number of triplets. Suppose there are $s$ keys (subjects) and $o$ values (corresponding amount of  object-based lists) in adjacency list. The theoretical logits amount and the estimated logits amount on two benchmark datasets (NYT and WebNLG) are shown in Table~\ref{tab:logit1_all}. From the viewpoint of predicted logits estimation, DIRECT is the most representative-efficient model.

%

\end{document}